\documentclass[runningheads]{llncs}

\usepackage{eccv}

\usepackage{eccvabbrv}

\usepackage{graphicx}
\usepackage{booktabs}

\usepackage[accsupp]{axessibility}  

\usepackage{hyperref}

\usepackage{orcidlink}
\usepackage{wrapfig}
\usepackage{booktabs}
\usepackage{multirow}
\usepackage{graphicx}

\begin{document}

\title{Online Learning via Memory: \\ Retrieval-Augmented Detector Adaptation} 


\author{Yanan Jian\thanks{Primary contributor.}\and
Fuxun Yu\thanks{Secondary contributor.}\and
Qi Zhang\and
William Levine\and
Brandon Dubbs\and
Nikolaos Karianakis}

\authorrunning{Y. Jian et al.}

\institute{Microsoft Corp\\
\email{yananjian@microsoft.com, fuxunyu@microsoft.com, qizhang5@microsoft.com, wlevine@microsoft.com, v-bdubbs@microsoft.com, nikolaos.karianakis@microsoft.com}\\}

\maketitle
\begin{abstract}
  This paper presents a novel way of online adapting any off-the-shelf object detection model to a novel domain without retraining the detector model. Inspired by how humans quickly learn knowledge of a new subject (e.g., memorization), we allow the detector to look up similar object concepts from memory during test time. This is achieved through a retrieval augmented classification (RAC) module together with a memory bank that can be flexibly updated with new domain knowledge. We experimented with various off-the-shelf open-set detector and close-set detectors. With only a tiny memory bank (e.g., 10 images per category) and being training-free, our online learning method could significantly outperform baselines in adapting a detector to novel domains.  
\end{abstract}
\section{Introduction}
\label{sec:intro}

Object detection has advanced significantly over the years. Traditionally, effective object detectors were developed by collecting a large amount of annotated training data and training the model offline, a process that could take days or weeks and result in models tightly coupled to specific datasets or domains, such as Faster-RCNN~\cite{fasterrcnn}, YOLO~\cite{yolo} and etc. on COCO dataset~\cite{cocodataset}. 
Recently, with advancements in foundational model-accelerated pseudo labeling pipelines and more comprehensive training datasets, object detectors have evolved from closed-set to open-world models. These new models, like GLIP~\cite{glip} and Grounding-DINO~\cite{g-dino}, are capable of detecting wide range of objects.

However, applying these models to novel domains often results in poor detection performance. For example, the state-of-the-art open-set model Grounding-DINO\cite{g-dino}, primarily trained on large-scale natural scene datasets, achieves \textbf{less than 1\% mAP} on the aerial dataset DOTA~\cite{dota}.
To adapt to new visual domains and class vocabularies, these models still require offline retraining or fine-tuning. Traditional offline learning paradigms face two major bottlenecks: (1) the need for extensive labeling of target-domain data to enable stable training, and (2) a time-consuming retraining/fine-tuning process. These overheads hinder the efficient adaptation of detector models in real-world scenarios.

%


%

\begin{figure}[!tb]
  \centering
  \vspace{-3mm}
  \includegraphics[width=12cm]{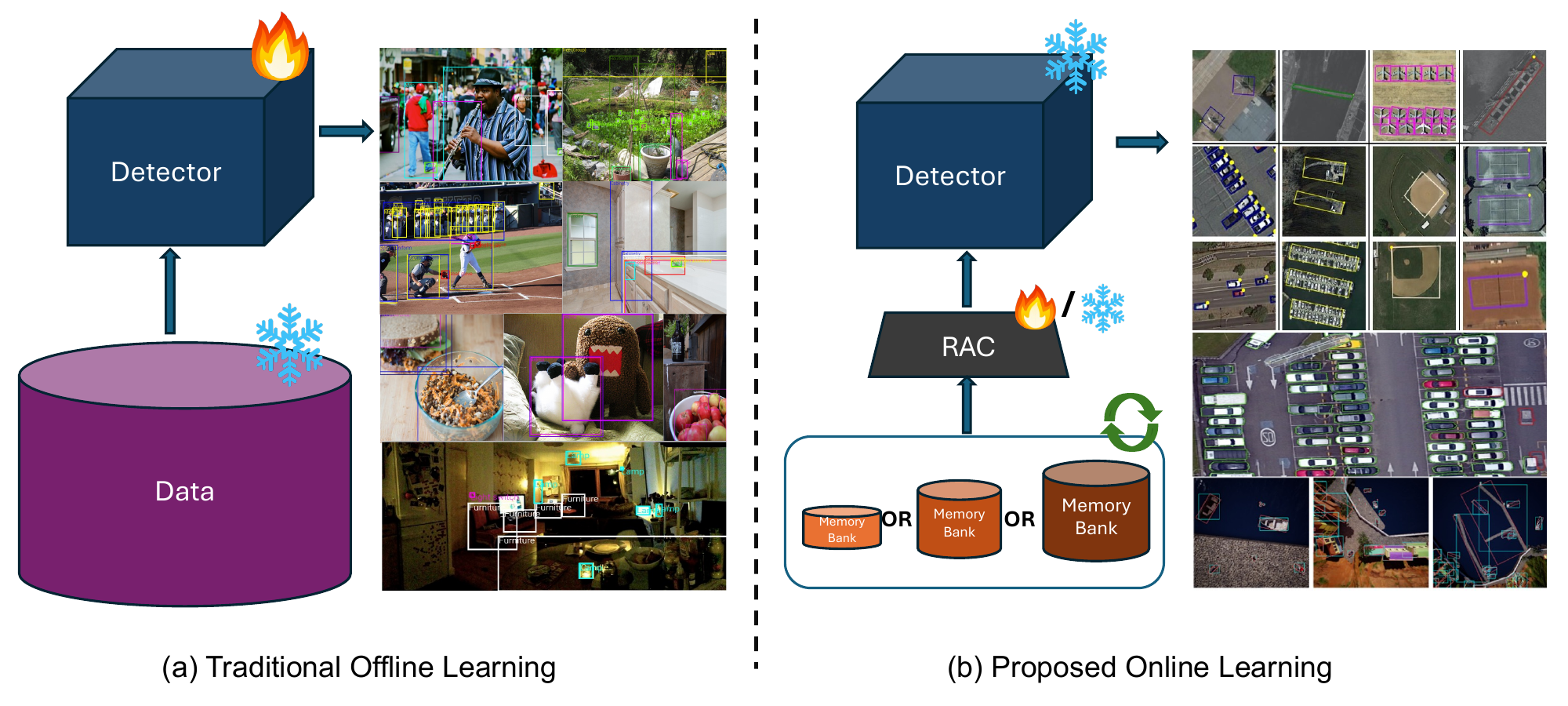}
  \vspace{-2mm}
  \caption{Comparison between (a) traditional \textbf{offline detector learning} vs. (b) \textbf{proposed online learning} paradigm. With an updatable memory bank and a retrieval-augmented (RAC) module, we could leverage any off-the-shelf frozen detector to quickly detect new domain concepts by retrieving them from memory bank.}
  \label{fig:offline_online_comparison}
    \vspace{-5mm}
\end{figure}

To address these challenges, we propose an innovative online learning framework that adapts any detector to new domains with minimal overhead.
The overview of our framework is shown in Fig.~\ref{fig:offline_online_comparison} (b).  Comparing to traditional offline learning in Fig.~\ref{fig:offline_online_comparison} (a) that requires detector re-training/finetuning, our method could freeze the off-the-shelf detector. Instead, we design two modules: (1) a \textbf{updatable memory bank} with a small set of label images in target domain; and (2) a \textbf{retrieval-augmented module} (RAC) to help classify new concepts by retrieving similar-apperance images and objects from the memory bank. 

Such an online learning paradigm provides following benefits: (1) it enables \textbf{rapid any-domain adaptation} as soon as minimal amounts of target-domain data (e.g., as less as 10 images per category) are labeled in the memory bank; (2) it allows the detector to conduct \textbf{online continual learning} for continuous improvement when more and more data are inferenced and/or labeled and added to the updatable memory bank; (3) all such capabilities comes \textbf{without any exra computing cost} in re-training/finetuning, which greatly facilites the real-world application deployment. 

The retrieval-augmented paradigm is relatively underexplored in online visual domain adaptation, but many of prior works has highlightened its potential in addressing visual online learning challenges. Notably, the retrieval-augmented paradigm has proven effective for large language models (LLMs) in addressing forgetting and hallucination in novel text domains (e.g., retrieval-augmented generation (RAG~\cite{rag}) and KNN-LM~\cite{knn-lm}). There are also existing works on retrieval-augmented visual models with three main categories: (1) Generalizing vision tasks by retrieving task-related images and tuning an additional task-specific encoder (e.g., REACT~\cite{learning-customized-visual-models-with-retrieval-augmented-knowledge}). (2) Solving long-tail classification-only problem by a huge augmented knowledge base~\cite{rac}. (3) Open-vocabulary object detection, which requires a complex multi-step training process and LLMs to function~\cite{retrieval-augmented-open-vocabulary-object-detection}.  

We construct an aerial domain adaptation benchmark with DOTA~\cite{dota} dataset. Experiments demonstrate the bottlenecks of pretrained close-set and open-set models (e.g., faster-rcnn, and grounding dino) in such novel domains. Without bells and whistles, our online learning method enables them to improve performance by a large margin and with minimal labeling efforts.




\section{Retrieval-Augmented Detector Adaptation}
\label{sec:method}

\begin{wrapfigure}{r}{0.55\textwidth}
  \centering
    \includegraphics[width=0.55\textwidth]{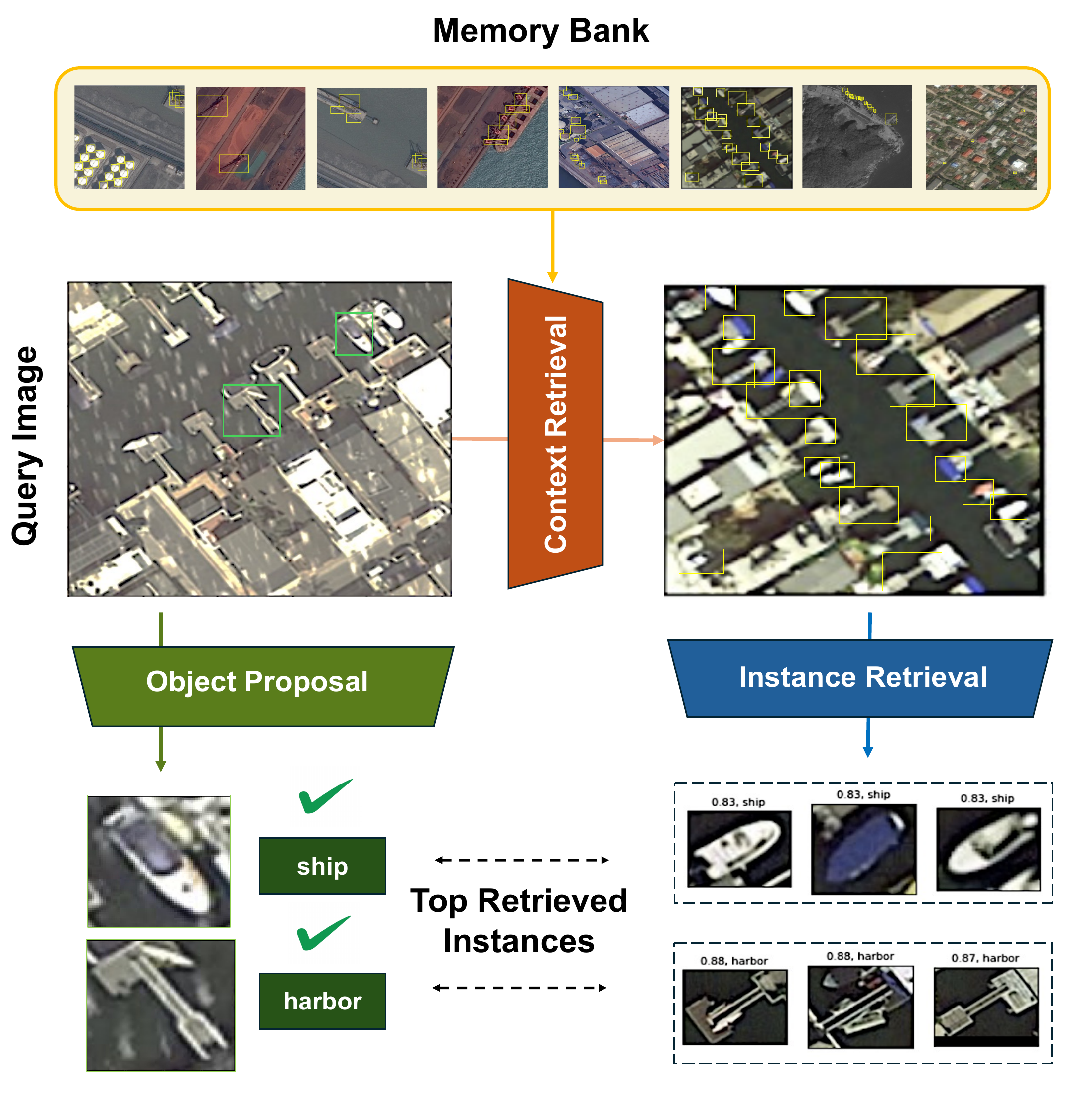}
  \vspace{-5mm}
  \caption{Context-Aware RAC Workflow.}
  \vspace{-7mm}
  \label{fig:rac_flow}
\end{wrapfigure}

\noindent \textbf{Overview~} As shown in Fig.~\ref{fig:rac_flow}, our proposed online learning framework is composed of the following major modules: 
(i) an online-updatable \textbf{memory bank}, which contains target-domain images that are used to supply new concepts for online adaptation; 
(ii) an \textbf{object proposal model} that is from off-the-shelf, 
either open-world detectors, or any detectors trained on similar domain data with different ontologies, or simply RPNs. 
(iii) a \textbf{context retrieval module} that can be used to associate the image contexts from memory bank with the inference images; 
and (iv) an \textbf{instance retrieval module} that associates proposed object instances with instances from similar context filtered by (iii). 
Fig. \ref{fig:rac_flow} shows the overall pipeline of online inference workflow with RAC: 
For a query image, context-level RAC first select top similar contextual images from memory bank. Then based on the object proposals from the query image, for each proposal, instance-level RAC performs instance matching constrained to the selected similar-context images. As a result, each proposal gets assigned a class based on votes from the top retrieved instances.



\subsection{Pre-requisite: Object Proposal Model}
Object detection tackles two key problems: localization and classification. In this paper, we adopt pre-trained detectors as object proposal networks for the localization subtask, and focus on adressing novel concept classification subtask. 

Therefore, our proposal networks could be of various forms, such as off-the-shelf open-set detectors, detectors trained on different dataset (e.g., with different ontology), or simply RPNs, as long as it could provide meaningful foreground proposals. 
For example, even with a binary RPN network without any semantic capability, our method could equip the binary RPN with classification capability to reach over $\sim$20mAP on 18-class DOTA benchmark. We will evaluate all aforementioned proposal networks (off-the-shelf grounding-DINO, pretrained RPN, and pretrained detector on a similar dataset) in the later experiments to demonstrate the effectiveness of our proposed framework.

\subsection{Memory Bank}

Our framework requires minimal data to construct the memory bank, such as 10 images per class, easily labeled by end-users in online learning settings. To construct an efficient memory bank, we propose an unsupervised image selection method using image-level feature clustering to maximize coverage and minimize labeling efforts. \par
\vspace{1mm}
\noindent \textbf{Unsupervised Seed Image Clustering~} We use a strong image feature extraction backbone (e.g. CLIP~\cite{clip}) to extract embeddings from unlabeled target domain images. These embeddings are then clustered (e.g., using k-means) into a target number of clusters based on the number of images end-users can label. The centroid images from each cluster, representing diverse and representative scenes, are annotated by end-users. This method allows us to achieve decent detection performance by labeling as few as 10 images per category.

\vspace{1mm}
\subsection{Retrieval-Augmented (RAC) Modules}

With the labeled seed objects and images in the memory bank, retrieval-augmented modules can enable object detectors to acquire new semantic classification capabilities by matching targeted detected proposals to seed objects.

One significant challenge in accurate object matching is the presence of objects from different categories with similar appearances in the target domain. This similarity can confuse the feature extractor and lead to inaccurate classification results in a training-free manner. For example, distinguishing between white small vehicles and small white boats in aerial images, or between container cranes and harbors, can be difficult. 

To address such confusions, we construct a multi-stage context matching process. The first stage, \textbf{context retrieval}, shrinks the search range by filtering out irrelevant scenes (e.g., filtering out maritime scenes for boats). The second stage, \textbf{instance retrieval}, is conducted from the context-matched images. By considering both instance appearance and context, this approach minimizes classification confusion and enhances retrieval accuracy.

For the retrieval-augmented model, a strong feature extractor is necessary. However, it does not need to be trained on the target domain to achieve decent semantic classification accuracy. Therefore, any strong pre-trained feature extractor, such as DINOV2~\cite{dinov2} or CLIP~\cite{clip}, can be used in a training-free manner or fine-tuned on the provided memory bank for best performance.

Specifically, for image-level semantic matching, we use the off-the-shelf CLIP model to extract image-level features and then calculate the similarity between the query image and memory bank images. For the second stage of instance-level matching, we select the top k images (k=20, 50, 100) from the image-level matching results, extract bounding box-level features using either the off-the-shelf or fine-tuned CLIP model, and then calculate instance-to-instance similarity, constrained to only the selected top k images. Consequently, the final instance classification results are an ensemble of both bounding box-level and global-context matching, effectively reducing appearance-induced confusion.


\section{Experiments}
\label{sec:exp}

\subsection{Experiment Setup}
\vspace{-2mm}
For novel domain dataset, we conduct online adaptation experiments on DOTA dataset. There are 18 classes in DOTA; besides a few classes that overlap with natural dataset concepts, e.g. plane, ship, large-vehicle, small-vehicle and helicopter, the other classes, e.g. storage-tank, container-crane, helipad are uncommon and have highly imbalanced class distribution. \footnote{Note that all class names are shown in the following tables as abbrevations. PL:plane, SP:ship, ST:storage-tank, BD:baseball-diamond, TC:tennis-court, BC:basketball-court, GTF:ground-track-field, HB:harbor, BG:bridge, LV:large-vehicle, SV:small-vehicle, HC:helicopter, RA:roundabout, SBF:soccer-ball-field, SW:swimming-pool, CC:container-crane, AP:airport, HP:helipad.}

\vspace{1mm}
\textbf{Object Proposal Network:} We evaluate three major types: (1) Grounding-DINO(\textit{`G-DINO'}): This state-of-the-art open-set model uses prompts composed of dot-connected, space-separated class names, as suggested by the authors ~\cite{g-dino}.
(2) xView-pretrained Faster-RCNN \cite{fasterrcnn}: This model employs a Swin-T backbone \cite{swin} and is pretrained on the xView dataset, which shares a similar visual domain with DOTA. But as its ontology does not overlap with DOTA, we removed all predicted labels and utilize the bounding boxes as proposals.
(3) DOTA-pretrained Faster-RCNN (RPN / ROI stages, bboxes without labels): This model will provide accurate bbox proposals without classification labels. Such an evaluation will give us insight into the upper-bound classification capability of RAC framework without being constrained by localization quality.

\vspace{1mm}
\textbf{Memory Bank:}  Memory bank of different selection will be ablated with two major settings:
(1) \textit{`Tiny DB'}: Sampled from the DOTA training set, with a maximum of 10 images per class, resulting in a total of 180 images. (2) \textit{`Base DB'}: Sampled from the DOTA training set, with a maximum of 250 images per class, resulting in a total of  $\sim$4,000 images. We will also evaluate intermediate settings of 1,000 - 2,000 images to demonstrate the impact of DB size.

\vspace{1mm}
\textbf{RAC Backbone:}  For the retrieval module, we will compare off-the-shelf CLIP ~\cite{clip} (\textit{`raw CLIP'}) and finetuned CLIP (\textit{`ft CLIP'}) trained on object instances from Tiny DB images in later experiments. Both are with ViT-B/32 structure. Detailed fine-tuning settings will be covered in subsequent subsections. 

\vspace{1mm}
\textbf{RAC Parameters} We use two thresholds for the RAC process: context-threshold and instance-threshold, tuned with a held-out set of 1000 images. Optimal results were achieved with {k=50 (number of retrieved images), n=1 (number of retrieved instances)}, context-thresh=0.1, and instance-thresh=0.8. Proposals with RAC scores below the threshold are discarded; those above are assigned the class ID of the top instance. The class score is calculated as $w1 \times proposal\_score + w2 \times instanceRAC\_cosine\_score$, with $w1 + w2 = 1.0$. For G-DINO proposals, $w1=0.2$ and $w2=0.8$; for others, $w1=0.5$ and $w2=0.5$.
Note that the params are only emperical values to demonstrate the effectiveness of our pipeine for simplicity. Better parameter tuning method can be designed for different scenarios, which we leave as future work.

\begin{table}[!tb]
\caption{Online Adaptation on DOTA Benchmark (GDINO and xView FasterRCNN)}
\vspace{-2mm}
\label{tab:my-table1}
\centering
\resizebox{0.75\textwidth}{!}{%
\renewcommand\arraystretch{1.3}
\setlength{\tabcolsep}{2.8mm}{
\begin{tabular}{l|ll|l|lllll}
\hline
\textbf{Model} & \multicolumn{2}{l|}{\textbf{Setting}} & \textbf{Mean AP} & \textbf{PL} & \textbf{SP} & \textbf{LV} & \textbf{SV} & \textbf{HC} \\ \hline
\multicolumn{1}{c|}{\multirow{5}{*}{G-Dino}} & \multicolumn{2}{l|}{Baseline} & 2.68 & 12.7 & 0.1 & 0.2 & 0.2 & 0.2 \\ \cline{2-9} 
\multicolumn{1}{c|}{} & \multicolumn{1}{c|}{\multirow{2}{*}{Raw CLIP}} & RAC (Tiny DB) & \textbf{3.72} & \textbf{16.1} & 0.5 & 0.2 & \textbf{1.0} & 0.8 \\
\multicolumn{1}{c|}{} & \multicolumn{1}{c|}{} & RAC (Base DB) & 3.7 & 15.7 & \textbf{0.7} & \textbf{0.7} & 0.3 & \textbf{1.1} \\ \cline{2-9} 
\multicolumn{1}{c|}{} & \multicolumn{1}{l|}{\multirow{2}{*}{Ft CLIP}} & RAC (Tiny DB) & \textbf{4.54} & \textbf{17.3} & \textbf{1.1} & \textbf{0.5} & 1.1 & \textbf{2.7} \\
\multicolumn{1}{c|}{} & \multicolumn{1}{l|}{} & RAC (Base DB) & 4.5 & 17.0 & 0.8 & \textbf{0.5} & \textbf{1.9} & 2.3 \\ \hline
\multirow{5}{*}{\begin{tabular}[c]{@{}l@{}}FasterRCNN\\  (XView, \\ w/o label)\end{tabular}} & \multicolumn{2}{l|}{Baseline} & \multicolumn{1}{c|}{-} & \multicolumn{1}{c}{-} & \multicolumn{1}{c}{-} & \multicolumn{1}{c}{-} & \multicolumn{1}{c}{-} & \multicolumn{1}{c}{-} \\ \cline{2-9} 
 & \multicolumn{1}{l|}{\multirow{2}{*}{Raw CLIP}} & RAC (Tiny DB) & 9.12 & 22.1 & 12.4 & 3.1 & 7.2 & 0.8 \\
 & \multicolumn{1}{l|}{} & RAC(Base DB) & \textbf{13.6} & \textbf{23.6} & \textbf{23.1} & \textbf{7.8} & \textbf{8.9} & \textbf{4.6} \\ \cline{2-9} 
 & \multicolumn{1}{l|}{\multirow{2}{*}{Ft CLIP}} & RAC (Tiny DB) & 11.34 & 24.3 & 15.1 & 4.8 & 6.8 & 5.7 \\
 & \multicolumn{1}{l|}{} & RAC (Base DB) & \textbf{14.7} & \textbf{25.1} & \textbf{23.4} & \textbf{9.1} & \textbf{9.0} & \textbf{6.9} \\ \hline
\end{tabular}%
}}
\vspace{1mm}
\end{table}

\subsection{Proposal Network Evaluation}

Table \ref{tab:my-table1} presents mean mAPs for five main classes on the DOTA validation set using G-DINO and the xView-pretrained detector (performance as zero since it's without classification labels). 
For G-DINO, we improve mAP from 2.68 to 3.72 (raw CLIP) and 4.54 (ft CLIP) with only 10 images per class. 
For xView-pretrained binary detector, RAC with raw CLIP could reach 13.6 mAP, and 14.7 mAP with ft CLIP. Especially, for airplane (PL) and ship (SP) classes, the performance of RAC could reach $\sim$25 and $\sim$23 AP.

\begin{table}[!tb]
\caption{Memory Bank Sampling Method: Random vs Unsupervised Clustering}
\vspace{-2mm}
\label{tab:sample_method}
\resizebox{\textwidth}{!}{%
\renewcommand\arraystretch{1.5}
\begin{tabular}{lll|ll|llllllllllllllllll}
\hline
\multicolumn{2}{l}{\textbf{Model}} & \textbf{Setting} & \textbf{mAP} & \textbf{mAR} & \textbf{PL} & \textbf{SP} & \textbf{ST} & \textbf{BD} & \textbf{TC} & \textbf{BC} & \textbf{GTF} & \textbf{HB} & \textbf{BG} & \textbf{LV} & \textbf{SV} & \textbf{HC} & \textbf{RA} & \textbf{SBF} & \textbf{SW} & \textbf{CC} & \textbf{AP} & \textbf{HP} \\ \hline
\multicolumn{1}{c|}{\multirow{3}{*}{G-Dino}} & \multicolumn{2}{l|}{Baseline} & 0.8 & 2.3 & 12.7 & 0.1 & 0 & 0 & 0 & 0 & 0 & 0 & 0 & 0.2 & 0.2 & 0.2 & 0 & 0 & 0.2 & 0 & 0 & 0 \\ \cline{2-23} 
\multicolumn{1}{c|}{} & \multicolumn{1}{l|}{\multirow{2}{*}{Raw CLIP}} & RAC + Rand-Tiny DB & 1.1 & 14.5 & 12.6 & 0.3 & \textbf{0.5} & 0.3 & 0 & 0 & 0.1 & 1.2 & 0 & 0.2 & \textbf{1.1} & 0.6 & 0.3 & 0.2 & \textbf{2.1} & 0 & 0.1 & \textbf{0.1} \\
\multicolumn{1}{c|}{} & \multicolumn{1}{l|}{} & RAC + Cluster-Tiny DB & \textbf{1.2} & \textbf{15.2} & \textbf{16.1} & \textbf{0.5} & 0.4 & 0.3 & 0 & 0 & 0.1 & \textbf{1.6} & 0 & 0.2 & 1 & \textbf{0.8} & \textbf{0.4} & 0.2 & 0.7 & 0 & 0.1 & 0 \\ \hline
\multicolumn{1}{l|}{\multirow{3}{*}{G-Dino}} & \multicolumn{2}{l|}{Baseline} & - & - & - & - & - & - & - & - & - & - & - & - & - & - & - & - & - & - &  &  \\ \cline{2-23} 
\multicolumn{1}{l|}{} & \multicolumn{1}{l|}{\multirow{2}{*}{Ft CLIP}} & RAC + Rand-Tiny DB & 1.7 & 19.1 & 15.4 & 0.7 & \textbf{0.9} & 0.4 & 0.2 & \textbf{0.4} & 0.3 & 2.9 & 0.7 & 0.4 & \textbf{1.3} & 1.2 & 1.4 & 0.7 & 3.9 & 0 & \textbf{0.4} & 0.2 \\
\multicolumn{1}{l|}{} & \multicolumn{1}{l|}{} & RAC + Cluster-Tiny DB & \textbf{2.1} & 19.1 & \textbf{17.3} & \textbf{1.1} & 0.4 & \textbf{0.5} & 0.2 & 0.1 & \textbf{0.6} & \textbf{3.1} & 0.7 & \textbf{0.5} & 1.1 & \textbf{2.7} & \textbf{3.7} & 0.7 & \textbf{4.3} & \textbf{0.1} & 0.3 & \textbf{0.3} \\ \hline
\end{tabular}%
}
\vspace{-2mm}
\end{table}

\vspace{-2mm}
\subsection{Memory Bank Ablation}

\paragraph{Memory Bank Sampling Method}

Next, we show the effectiveness of our unsupervised clustering-based sampling method of constructing the memory bank. Specifically for Tiny DB, we comapre random sampling and cluster sampling. For cluster sampling, we use K-means with 10 clusters for each class and then randomly select one image per cluster, per class. 

As shown in Table \ref{tab:sample_method}, the cluster-sampled Tiny DB outperforms the random-sampled Tiny DB in both light-training and training-free scenarios, likely due to the increased diversity ensured by cluster-sampling.
Based on these results, we choose the better-performing cluster-sampled Tiny DB for all experiments involving Tiny DB.

\vspace{-2mm}
\paragraph{Memory Bank Size Comparison}

We then apply base DB to compare performance. The results show significant AP lifts in both training-free (Table \ref{tab:dbsize_train_free}) and light-training (Table \ref{tab:dbsize_train_light}) settings with the increased DB size, especially with better localization quality (comparing RPN vs ROI stages as object proposal network). 

Notably, we find the quality of object proposals significantly affects the performance of RAC. For example, the ROI stage from FasterRCNN trained on DOTA provides the most accurate bounding box proposals. With these proposals, purely relying on RAC for classification, the overall framework could achieves the highest 27.4 mAP on DOTA benchmark.

\begin{table}[!t]
\caption{Tiny DB vs. Base DB, training free}
\vspace{-2mm}
\label{tab:dbsize_train_free}
\resizebox{\textwidth}{!}{%
\renewcommand\arraystretch{1.5}
\setlength{\tabcolsep}{0.8mm}{
\begin{tabular}{ll|ll|llllllllllllllllll}
\hline
\textbf{Model} & \textbf{Setting} & \textbf{mAP} & \textbf{mAR} & \textbf{PL} & \textbf{SP} & \textbf{ST} & \textbf{BD} & \textbf{TC} & \textbf{BC} & \textbf{GTF} & \textbf{HB} & \textbf{BG} & \textbf{LV} & \textbf{SV} & \textbf{HC} & \textbf{RA} & \textbf{SBF} & \textbf{SW} & \textbf{CC} & \textbf{AP} & \textbf{HP} \\ \hline
\multicolumn{1}{c}{\multirow{3}{*}{G-Dino}} & Baseline & 0.8 & 2.3 & 12.7 & 0.1 & 0 & 0 & 0 & 0 & 0 & 0 & 0 & 0.2 & 0.2 & 0.2 & 0 & 0 & 0.2 & 0 & 0 & 0 \\ \cline{2-22} 
\multicolumn{1}{c}{} & RAC + Tiny DB & 1.2 & \textbf{15.2} & 16.1 & \textbf{0.5} & 0.4 & 0.3 & 0 & 0 & 0.1 & 1.6 & 0 & 0.2 & 1 & 0.8 & 0.4 & 0.2 & 0.7 & 0 & \textbf{0.1} & \textbf{0} \\
\multicolumn{1}{c}{} & RAC + Base DB & \textbf{1.5} & \textbf{15.2} & \textbf{16.4} & 0.4 & \textbf{0.9} & \textbf{0.4} & \textbf{0.1} & \textbf{0.1} & \textbf{0.6} & \textbf{2.1} & 0 & \textbf{0.3} & \textbf{1.3} & \textbf{1.2} & \textbf{0.8} & \textbf{0.6} & \textbf{1.7} & 0 & \textbf{0.1} & \textbf{0.5} \\ \hline
\multirow{3}{*}{RPN (DOTA)} & Baseline & - & - & - & - & - & - & - & - & - & - & - & - & - & - & - & - & - & - & - & - \\ \cline{2-22} 
 & RAC + Tiny DB & 6.3 & 24.8 & 35.5 & 12.4 & 6.6 & 5.6 & 16.5 & 1.7 & 2.6 & 5.9 & 1 & 5.3 & 7.3 & 4.4 & 2.9 & 1.8 & 3.4 & 0 & 1.1 & \textbf{0.2} \\
 & RAC + Base DB & \textbf{13.1} & \textbf{25.5} & \textbf{39.7} & \textbf{29.1} & \textbf{16.7} & \textbf{10} & \textbf{48.3} & \textbf{6.9} & \textbf{4.6} & \textbf{13.4} & \textbf{5.4} & \textbf{19.2} & \textbf{11} & \textbf{11.7} & \textbf{6.8} & \textbf{2} & \textbf{11} & 0 & \textbf{3} & 0 \\ \hline
\multirow{3}{*}{ROI (DOTA)} & Baseline & - & - & - & - & - & - & - & - & - & - & - & - & - & - & - & - & - & - & - & - \\ \cline{2-22} 
 & RAC + Tiny DB & 12.7 & 24.3 & 48 & 23.8 & 22.2 & 11.6 & 54 & 6.5 & 4 & 10.2 & 1.3 & 12.4 & 14.4 & 1.8 & 5.2 & 2.9 & 6.1 & 0 & 5 & 0 \\
 & RAC + Base DB & \textbf{22.0} & \textbf{28.5} & \textbf{53.9} & \textbf{46} & \textbf{29.8} & \textbf{17.9} & \textbf{68.5} & \textbf{19.1} & \textbf{13.2} & \textbf{29.9} & \textbf{6.7} & \textbf{35.5} & \textbf{20.2} & \textbf{10.9} & \textbf{8.3} & \textbf{9.1} & \textbf{14.8} & 0 & \textbf{11.6} & 0 \\ \hline
\end{tabular}%
}}
\vspace{2mm}
\end{table}

\begin{table}[!tb]
\vspace{-2mm}
\caption{Tiny DB vs. Base DB, with light training}
\vspace{-1mm}
\label{tab:dbsize_train_light}
\resizebox{\textwidth}{!}{%
\renewcommand\arraystretch{1.5}
\setlength{\tabcolsep}{0.8mm}{
\begin{tabular}{ll|ll|llllllllllllllllll}
\hline
\textbf{Model} & \textbf{Setting} & \textbf{mAP} & \textbf{mAR} & \textbf{PL} & \textbf{SP} & \textbf{ST} & \textbf{BD} & \textbf{TC} & \textbf{BC} & \textbf{GTF} & \textbf{HB} & \textbf{BG} & \textbf{LV} & \textbf{SV} & \textbf{HC} & \textbf{RA} & \textbf{SBF} & \textbf{SW} & \textbf{CC} & \textbf{AP} & \textbf{HP} \\ \hline
\multirow{3}{*}{G-Dino} & Baseline & 0.8 & 2.3 & 12.7 & 0.1 & 0 & 0 & 0 & 0 & 0 & 0 & 0 & 0.2 & 0.2 & 0.2 & 0 & 0 & 0.2 & 0 & 0 & 0 \\ \cline{2-22} 
 & RAC + Tiny DB & 2.1 & \textbf{19.1} & \textbf{17.3} & \textbf{1.1} & 0.4 & 0.5 & 0.2 & 0.1 & 0.6 & 3.1 & 0.7 & 0.5 & 1.1 & \textbf{2.7} & \textbf{3.7} & 0.7 & 4.3 & 0.1 & 0.3 & \textbf{0.3} \\
 & RAC + Base DB & \textbf{2.4} & 16.3 & 17 & 0.8 & \textbf{0.9} & \textbf{1} & \textbf{0.5} & \textbf{0.8} & \textbf{1} & \textbf{4.4} & 0.7 & 0.5 & \textbf{1.9} & 2.3 & 3.1 & \textbf{1.4} & \textbf{5.6} & \textbf{0.8} & \textbf{0.5} & 0 \\ \hline
\multirow{3}{*}{RPN (DOTA)} & Baseline & - & - & - & - & - & - & - & - & - & - & - & - & - & - & - & - & - & - & - & - \\ \cline{2-22} 
 & RAC + Tiny DB & 14.6 & \textbf{30.8} & 42.7 & 17.6 & 12.2 & \textbf{18} & 54.2 & 13.2 & \textbf{17.7} & 8.3 & 3.5 & 12.6 & 8 & 11.7 & \textbf{16.3} & \textbf{6.9} & 16.2 & 0 & 3.4 & \textbf{0.9} \\
 & RAC + Base DB & \textbf{17.7} & 29.5 & \textbf{45.6} & \textbf{32.9} & \textbf{21.4} & 17.9 & \textbf{57.7} & \textbf{22} & 13 & \textbf{13.1} & \textbf{5.9} & \textbf{25} & \textbf{11.1} & \textbf{15.5} & 13.6 & 4.3 & \textbf{17.5} & 0 & \textbf{4.1} & 0 \\ \hline
\multirow{3}{*}{ROI (DOTA)} & Baseline & - & - & - & - & - & - & - & - & - & - & - & - & - & - & - & - & - & - & - & - \\ \cline{2-22} 
 & RAC + Tiny DB & 21.8 & 34.4 & 56 & 31.7 & 27.9 & \textbf{27.7} & 71.8 & 22.2 & \textbf{28.3} & 17.4 & 5.4 & 22.9 & 16.7 & 7.5 & 15.1 & 13.5 & 16.9 & 0 & 9.3 & \textbf{1.5} \\
 & RAC + Base DB & \textbf{27.4} & \textbf{34.7} & \textbf{57.2} & \textbf{47.5} & \textbf{33.4} & 27.2 & \textbf{73} & \textbf{27} & 27.7 & \textbf{34.2} & \textbf{11} & \textbf{40.2} & \textbf{21.1} & \textbf{20.1} & \textbf{17.2} & \textbf{17.2} & \textbf{22.5} & 0 & \textbf{17.1} & 0 \\ \hline
\end{tabular}%
}}
\end{table}

\begin{figure}[!tb]
  \centering
  \includegraphics[width=\textwidth]{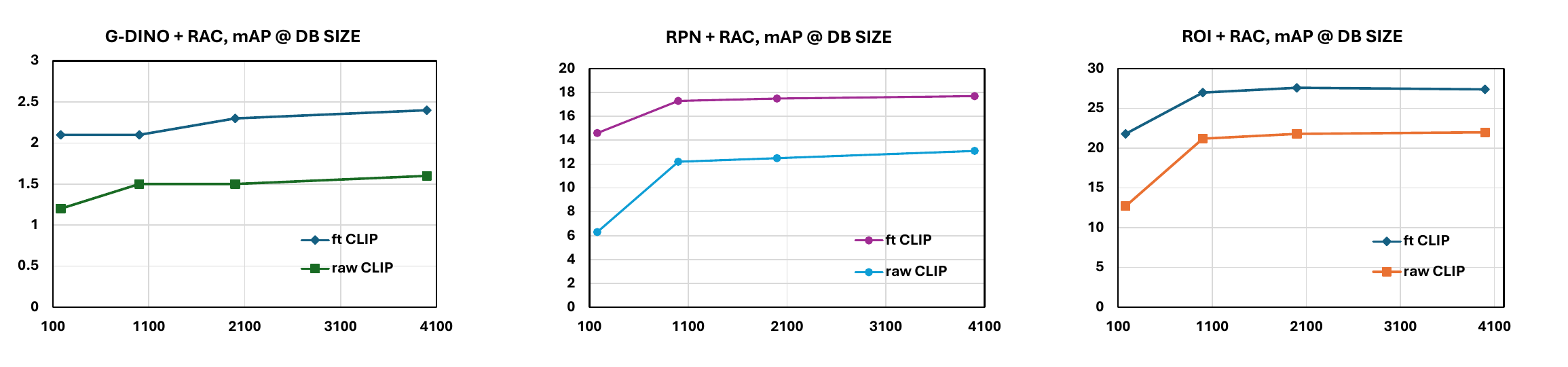}
    \vspace{-7mm}
  \caption{Mean AP vs Memory Bank Size Ablation.}
  \vspace{-7mm}
  \label{fig:dbsize}
\end{figure}

We further ablate the impact of DB size on mAP of the aforementioned object proposal models in a more fine-grained manner. As shown in Fig.~\ref{fig:dbsize}, for high-quality bounding box proposals (ROI), mAP reaches the highest $\sim$25 mAP at $\sim$1100 images in the DB (roughly $\sim$60 images per category), which demonstrates the efficiency of the RAC online learning method.

\vspace{-3mm}
\subsection{RAC Backbone Ablation: Pre-trained vs Finetuning}

Table \ref{tab:raw_vs_ft} shows that very light-weight fine-tuning CLIP on the tiny DB images enhances performance across all models. We use raw CLIP as the context RAC backbone, and the same CLIP model (raw or finetuned) for instance RAC matching. When fine-tuning, we train it with up to 500 object instances from Tiny DB for 5 epochs at a learning rate of 5e-4, a process that takes less than an hour on one V100 GPU. As the table shows, even without fine-tuning, the RAC method also significantly outperforms the baseline. Fine-tuning further boosts AP/AR for all classes, especially with high-quality proposals like with ROI stage.


\begin{table}[!tb]
\vspace{-2mm}
\caption{training-free vs. light-training, given Tiny DB}
\vspace{-2mm}
\label{tab:raw_vs_ft}
\resizebox{\textwidth}{!}{%
\renewcommand\arraystretch{1.5}
\setlength{\tabcolsep}{0.8mm}{
\begin{tabular}{ll|ll|llllllllllllllllll}
\hline
\textbf{Model} & \textbf{Setting} & \textbf{mAP} & \textbf{mAR} & \textbf{PL} & \textbf{SP} & \textbf{ST} & \textbf{BD} & \textbf{TC} & \textbf{BC} & \textbf{GTF} & \textbf{HB} & \textbf{BG} & \textbf{LV} & \textbf{SV} & \textbf{HC} & \textbf{RA} & \textbf{SBF} & \textbf{SW} & \textbf{CC} & \textbf{AP} & \textbf{HP} \\ \hline
\multirow{3}{*}{G-Dino} & Baseline & 0.8 & 2.3 & 12.7 & 0.1 & 0 & 0 & 0 & 0 & 0 & 0 & 0 & 0.2 & 0.2 & 0.2 & 0 & 0 & 0.2 & 0 & 0 & 0 \\ \cline{2-22} 
 & RAC + Raw CLIP & 1.2 & 15.2 & 16.1 & 0.5 & 0.4 & 0.3 & 0 & 0 & 0.1 & 1.6 & 0 & 0.2 & 1 & 0.8 & 0.4 & 0.2 & 0.7 & 0 & 0.1 & 0 \\
 & RAC + Ft CLIP & \textbf{2.1} & \textbf{19.1} & \textbf{17.3} & \textbf{1.1} & 0.4 & \textbf{0.5} & \textbf{0.2} & \textbf{0.1} & \textbf{0.6} & \textbf{3.1} & \textbf{0.7} & \textbf{0.5} & \textbf{1.1} & \textbf{2.7} & \textbf{3.7} & \textbf{0.7} & \textbf{4.3} & \textbf{0.1} & \textbf{0.3} & \textbf{0.3} \\ \hline
\multirow{3}{*}{RPN (DOTA)} & Baseline & - & - & - & - & - & - & - & - & - & - & - & - & - & - & - & - & - & - & - & - \\ \cline{2-22} 
 & RAC + Raw CLIP & 6.3 & 24.8 & 35.5 & 12.4 & 6.6 & 5.6 & 16.5 & 1.7 & 2.6 & 5.9 & 1 & 5.3 & 7.3 & 4.4 & 2.9 & 1.8 & 3.4 & 0 & 1.1 & 0.2 \\
 & RAC + Ft CLIP & \textbf{14.6} & \textbf{30.8} & \textbf{42.7} & \textbf{17.6} & \textbf{12.2} & \textbf{18} & \textbf{54.2} & \textbf{13.2} & \textbf{17.7} & \textbf{8.3} & \textbf{3.5} & \textbf{12.6} & \textbf{8} & \textbf{11.7} & \textbf{16.3} & \textbf{6.9} & \textbf{16.2} & 0 & \textbf{3.4} & \textbf{0.9} \\ \hline
\multirow{3}{*}{ROI (DOTA)} & Baseline & - & - & - & - & - & - & - & - & - & - & - & - & - & - & - & - & - & - & - & - \\ \cline{2-22} 
 & RAC + Raw CLIP & 12.7 & 24.3 & 48 & 23.8 & 22.2 & 11.6 & 54 & 6.5 & 4 & 10.2 & 1.3 & 12.4 & 14.4 & 1.8 & 5.2 & 2.9 & 6.1 & 0 & 5 & 0 \\
 & RAC + Ft CLIP & \textbf{21.8} & \textbf{34.4} & \textbf{56} & \textbf{31.7} & \textbf{27.9} & \textbf{27.7} & \textbf{71.8} & \textbf{22.2} & \textbf{28.3} & \textbf{17.4} & \textbf{5.4} & \textbf{22.9} & \textbf{16.7} & \textbf{7.5} & \textbf{15.1} & \textbf{13.5} & \textbf{16.9} & 0 & \textbf{9.3} & \textbf{1.5} \\ \hline
\end{tabular}%
}}
\vspace{-3mm}
\end{table}

\vspace{-2mm}
\subsection{Visual Analysis for RAC Process}

\begin{figure}[tb]
  \centering
  \includegraphics[width=0.9\textwidth]{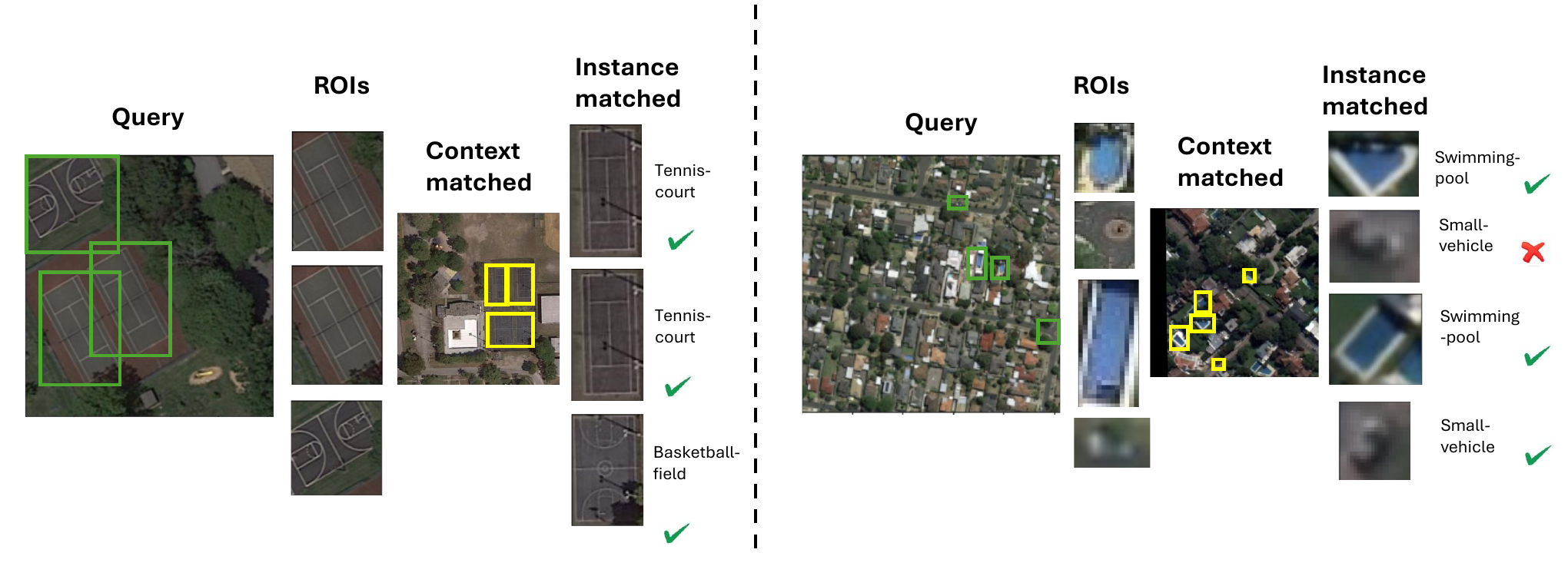}
    \vspace{-2mm}
  \caption{Visual Analysis for the RAC Process.
  }
  \vspace{-5mm}
  \label{fig:more_examples}
\end{figure}

Finally, we conduct visual analysis for the context-aware RAC process. Figure \ref{fig:more_examples} shows our context-aware RAC results in each intermediate step. 

Specifically, the green boxes are the proposed objects by the proposal network, and the yellow boxes are the top matched instances on the context-matched ground-truth images retrieved from the memory bank. As we can see, the RAC process through context-aware retrieval could largely shrink the search space and thus reduce the possibilities of misclassification, which leads to more accurate instance classification results.

\vspace{-3mm}
\section{Conclusion}
\label{sec:intro}

We propose a novel online learning framework through a retrieval-augmented classification process. 
Compared to traditional offline-training/fine-tuning based methods, our framework features \textbf{(1) online and continual learning capability}, \textbf{(2) minimal labeling requirement}, and \textbf{(3) no computing requirement} for visual domain adaptation. 
Our experiments demonstrate that the proposed method of RAC with tiny memory bank (minimal size of 10 images per category) enables fast and flexible online adaptation for detectors like grounding-dino and pretrained faster-rcnns on novel domain imagery benchmark. Notably, adaptation performance is optimal when object proposals are accurate. Therefore, a strong `detect-anything' model is also vital for visual object detector domain adaptation, which we leave as future work. 



%
%
\bibliographystyle{splncs04}
\bibliography{main}
\end{document}